\documentclass[letterpaper, 10 pt, conference]{ieeeconf}  

\IEEEoverridecommandlockouts                              

\usepackage{graphics} 
\usepackage{epsfig} 
\usepackage{mathptmx} 
\usepackage{times} 
\usepackage{amsmath} 
\usepackage{amssymb}  
\usepackage{multirow}
\usepackage{hhline}
\usepackage{color,soul}
\title{\LARGE \bf
Multi-Fingered In-Hand Manipulation with Various Object Properties \\
Using Graph Convolutional Networks and Distributed Tactile Sensors
}

\author{Satoshi Funabashi, Tomoki Isobe, Fei Hongyi, Atsumu Hiramoto, \\ Alexander Schmitz, Shigeki Sugano and Tetsuya Ogata
\thanks{The authors are with Waseda University, Okubo 3-4-1, Shinjuku, Tokyo 169-8555, Japan. (e-mail: s-funabashi@aoni.waseda.jp).}
}

\begin{document}

\maketitle
\thispagestyle{empty}
\pagestyle{empty}

\begin{abstract}

Multi-fingered hands could be used to achieve many dexterous manipulation tasks, similarly to humans, and tactile sensing could enhance the manipulation stability for a variety of objects. However, tactile sensors on multi-fingered hands have a variety of sizes and shapes. Convolutional neural networks (CNN) can be useful for processing tactile information, but the information from multi-fingered hands needs an arbitrary pre-processing, as CNNs require a rectangularly shaped input, which may lead to unstable results. Therefore, how to process such complex shaped tactile information and utilize it for achieving manipulation skills is still an open issue. This paper presents a control method based on a graph convolutional network (GCN) which extracts geodesical features from the tactile data with complicated sensor alignments. Moreover, object property labels are provided to the GCN to adjust in-hand manipulation motions. Distributed tri-axial tactile sensors are mounted on the fingertips, finger phalanges and palm of an Allegro hand, resulting in 1152 tactile measurements. Training data is collected with a data-glove to transfer human dexterous manipulation directly to the robot hand. The GCN achieved high success rates for in-hand manipulation. We also confirmed that fragile objects were deformed less when correct object labels were provided to the GCN. When visualizing the activation of the GCN with a PCA, we verified that the network acquired geodesical features. Our method achieved stable manipulation even when an experimenter pulled a grasped object and for untrained objects. A Project page including accompanying video and supplementary materials can be found at https://sites.google.com/site/bashifunabashi/multi-finger-project/in-hand-manipulation
\end{abstract}

\section{INTRODUCTION}

Humans use their multi-fingered hands for dexterous manipulation. Fingers moving in synchrony realize various manipulation skills such as grasping (e.g. power/precision grasps), in-hand manipulation including rolling contact, and finger gaiting \cite{manipgrasp}. Furthermore, tactile perception by human skin supports those manipulations \cite{codingsignal}. Also when it comes to robotic manipulation, tactile sensing is getting popular and is used for many manipulation tasks these days \cite{recenttacviz}. 
In our previous work we also established that 3-axis tactile sensing is beneficial compared to 1-axis sensors for multi-fingered tasks \cite{funabashiicra}. Furthermore, as contacts can occur with various parts of the robot hand, it is advantageous if all relevant parts of the hand are covered with skin sensors, not only the fingertips. 
Moreover, the physical properties of the manipulated objects should be considered to avoid dropping or deforming them; this is still a challenging issue \cite{trendchallnge}. 

In our past work, first, two-fingered manipulation was achieved using convolutional neural networks (CNNs) \cite{funabashiiros2020}. CNNs were used as they can process spatially distributed information. We also confirmed that one CNN could produce various in-grasp manipulation motions using labels allocated for each motion \cite{funabashiiros20202}.
Moreover, combined CNNs were proposed and used for object recognition with a multi-fingered hand \cite{funabashiicra}. In the study, the tactile sensors on the fingertips and phalanges have different sizes and shapes and the number of sensors mounted on each finger was different. 
   \begin{figure}[t]
      \centering
      \includegraphics[scale=0.32]{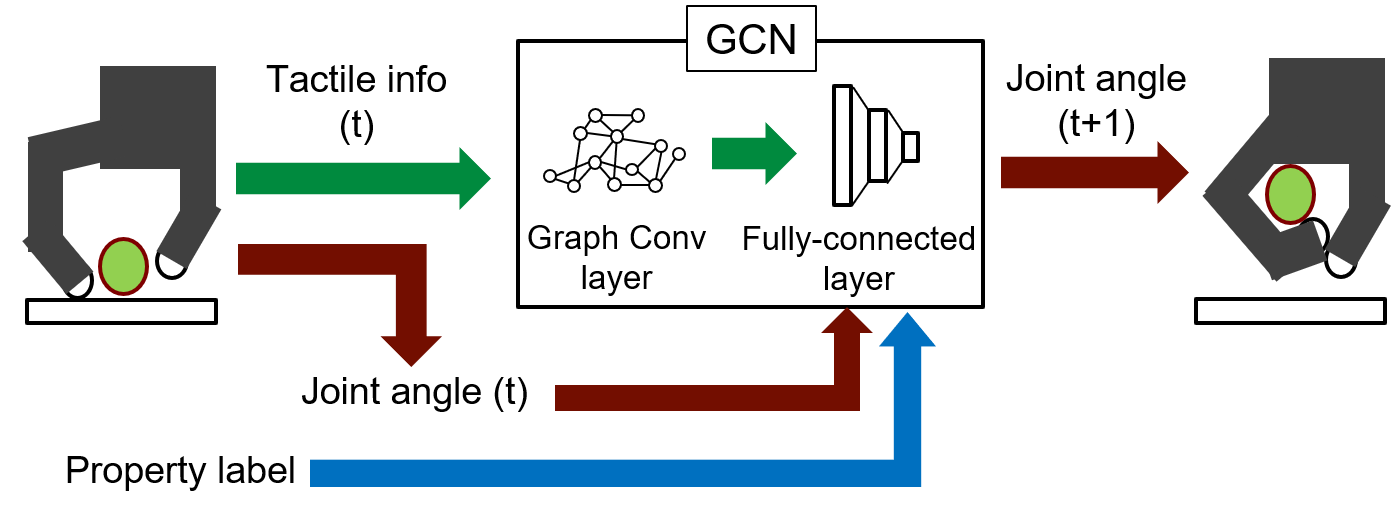}
      \caption{Schematic of the proposed motion-generating method.}
      \label{schematic}
   \end{figure}

However, the CNNs in \cite{funabashiicra} have a crucial problem. Sensors that are close to each other cannot always retain their neighborhood as input for the CNN, especially for curved fingertips; on the other hand, sensors that are not close to each other, such as from different fingers, are mapped next to each other. Moreover, the CNNs convolute tactile features always in the same order of fingers, thus for example the tactile information from the thumb is always next to that of the index finger. This does not always reflect the real robot configuration during in-hand manipulation as the thumb sometimes is closer to other fingers. Furthermore, the input always needs to be converted to a rectangular shape for the CNNs.
Overall, the experimenter will need to map the input from the tactile sensors to a form that is suitable for a CNN, thereby necessarily deforming the tactile information. The chosen mapping might not be appropriate for the task at hand and the result will vulnerably change according to the mapping chosen by the experimenter.



For these reasons, this study employs a graph convolutional network (GCN) which can 
map the relative positions of the sensors in a manner which more closely reflects the real configuration of the sensors on the hand.
The GCN is commonly used for graph-structured applications such as molecules and traffic networks. GCNs have been also applied to tactile sensing \cite{tactile-gcn}. However, the GCNs were only applied to an area on the fingertips and did not consider the structure of the robot hand. Our study investigates a GCN for tactile sensor alignments by following the robotic hand's configuration geodesically. 
Moreover, object property labels for each target object were prepared to adjust the manipulation motion so that one GCN manipulates various objects and does not break or drop the objects. Fig. \ref{schematic} shows a schematic of the proposed method.

Therefore, this paper presents these contributions: 
\begin{itemize}
\item GCN applied to intricate tactile sensor alignments on a multi-fingered hand.
\item Object property labels for adjusting in-hand manipulation motions.
\item Robust in-hand manipulation against a disturbance on the grasped object by an experimenter.
\item In-hand manipulation with unknown objects using the GCN and the distributed 3-axis tactile sensors.
\end{itemize}

\section{Related Work}
\subsection{Control System for Multi-Fingered In-Hand Manipulation}
Many multi-fingered hand control systems have been developed so far. Modeling and optimization methods were used for precise manipulation and object posing \cite{localplan}\cite{relaxedregidity}. Those methods usually focus on precise manipulation as they target precision grasps. However, their methods are difficult to extend to diverse manipulation motions of multi-fingered hands as they build kinematic contact models between the fingertips and a grasped object which are not applicable to other parts of the hands or manipulation tasks. In addition, multi-fingered manipulation sometimes requires a dynamic change of grasping states to change the object orientation or position in the hand. For example, to change the orientation of an object, force and friction between the fingertips and the object should be reduced which makes the contact area on the object small. This lets the hand re-orient the object easily, while it also means grasping states will be unstable when finger-gaiting or rolling contacts happen. To change between stable and unstable grasping states, a controller needs to recognize tactile events in the grasping states on each part of the hand at the same time. \cite{relaxedregidity} had a key insight to relax rigidity constraints between fingertips and a grasped object to change the position of an object. However, when to relax or constrain the grasping states is difficult to decide by checking the grasping states on the multi-fingered hands entirely as those methods have the aforementioned limitations.

Reinforcement learning achieved such dynamic manipulations with numerous contacts on an entire hand. Famous research achieved dexterous Rubik’s Cube multi-fingered manipulation \cite{DBLP:journals/corr/abs-1808-00177}. A limitation was that the palm of the hand always prevented the cube from dropping which made the task easier for the fingers. Even though the dexterous motion was acquired autonomously, it required many cameras for training and inference of the manipulation and had a huge training cost and hardware load. Learning from tactile sensors was also mentioned but not used in the end as the sensor information was difficult to model in their simulator. Another study used tactile information to achieve a variety of manipulations. However, it was only performed in simulation \cite{multifinsim}. Although those control methods skillfully achieved in-hand manipulation including dexterous motions such as finger-gaiting and/or rolling contact, they were not applied to a variety of objects in the real world \cite{reinforcemultilowcost}. In addition, despite the fact that tactile information is crucial for multi-fingered manipulation, it is still difficult to use for the manipulation of various objects.

While reinforcement learning is a powerful tool, it has the aforementioned limitations. The learning process can be accelerated by providing human motion data via tele-operation systems \cite{nvidiatele}, which can produce natural training data for deep learning based methods. 
However, in previous research the learned skills were not generalized to a variety of objects. 
One challenge 
can be that the manipulation motions with objects for multi-fingered hands are different depending on the physical properties of the objects such as size, shape, softness and slipperiness. Labels for the motions can be used to achieve adaptive motions with one controller. For example, in \cite{onehotsergey} one robotic controller achieved several motions. However, those control methods did not use a lot of tactile sensors, which provide information about the object size and shape for example. Therefore, tactile sensors have the potential for enabling robust in-hand manipulation with various objects. How to process tactile information for dexterous multi-fingered in-hand manipulation is still an open issue.
 
\subsection{Object Property and Processing Tactile Information}
Tactile sensors enable robotic hands to do dexterous manipulation with a variety of objects. Specifically, tactile sensors for fingertips such as Biotac \cite{slip1} and GelSight \cite{shape} are widely used. Using these kinds of tactile sensors are beneficial for achieving not only manipulation but also object recognition or detection of object-related events such as slip \cite{slipmulti}. Tactile exploration is a good way to effectively acquire such object information \cite{optrigsof}\cite{simuexp}. Many multi-fingered hands with such tactile sensors were developed \cite{towardsofthand}\cite{branden} but they focused on only fingertips. 

Most previous tactile sensors could not cover other parts of the hand with tactile sensors.  
Furthermore, according to our previous work \cite{funabashiicra}, 3-axis tactile sensors provide more useful information compared to 1-axis sensors. Fingertips with a human-like shape are beneficial for in-hand manipulation \cite{or2016}. 
To fulfil these requirements, we cover the fingertips, phalanges and palm of an Allegro hand with uSkin tactile sensors, in a similar configuration to our previous works \cite{funabashiicra}\cite{funabashiiros2020}.

For processing abundant tactile information to extract object information, CNNs have been widely used for tactile based robotic tasks \cite{clothtouch}. Since different sized and shaped tactile sensors are mounted on multi-fingered hands and CNNs require an input in a rectangular shape, cascaded CNNs were proposed \cite{funabashiicra}. This enabled to process the tactile information according to the morphology of the hand, but tactile information needed to be reshaped and fused in a subjective manner. 

Therefore, our current work uses a GCN, which was introduced by \cite{gcn}. A GCN has been used for a Biotac sensor which has irregularly placed tactile sensors \cite{tactile-gcn}. Moreover, a GCN was applied to the joint configuration of a multi-fingered hand \cite{livinggraph}, but neighboring fingers were connected to each other. As discussed before, the relative finger position changes during manipulation, and \cite{livinggraph} therefore has a problem similarly to our previous work \cite{funabashiicra}. 
Overall, GCNs have not been applied to tactile sensors following the robotic configuration, yet.

\section{Proposed Method}
\subsection{Allegro Hand with Tactile Sensors}
   
An Allegro Hand, a multi-fingered robotic hand made by Wonik Robotics, was used in this study. Each finger has 4 DOFs (16 DOFs in total). The uSkin distributed tactile sensors, which were used in our previous study\cite{funabashiicra}\cite{funabashiiros2020}, use 3-axis Hall effect magnetometers and small magnets located above them embedded in soft material. When forces are applied to the skin, the soft layer deforms, and accordingly the magnetic field. Considering that this research focuses on in-hand manipulation which has a contact on a vast area of the surface of the hand, tactile sensors are mounted on the fingertips, finger phalanges and palm as shown in Fig. \ref{sensormap}. 
Overall, the following measurements are used: 16 (4 fingers * 4 joint angles) + 1152 ((4 fingertips * 24 uSkin sensor chips) + (11 finger phalanges + 7 uSkin sensors on a palm) * 16 uSkin sensor chips) * 3 axes = 1168 measurements. Those sensor information is collected at a speed of 100 Hz.

\subsection{GCN for Geodesical Tactile Mapping}

A convolutional neural network (CNN) is a popular network used for processing tactile information spatially. Since the Allegro Hand manipulates objects using multiple fingers and a palm in this study, the CNN needs to receive tactile information from those parts of the hand. However, it is difficult to combine the tactile information of all fingers and the palm into a single map. This is a crucial problem of a CNN and thus a CNN is not used in this study. Therefore, a GCN was introduced as an alternative network to CNN, which could still consider tactile information spatially. Recurrent neural networks including long short term memory (LSTM) were not used as this study focused on geometric or geodesical tactile information even though the networks are useful for tasks with time-series information including multi-fingered manipulation.

\begin{figure}[t]
      \centering
      \includegraphics[scale=0.47]{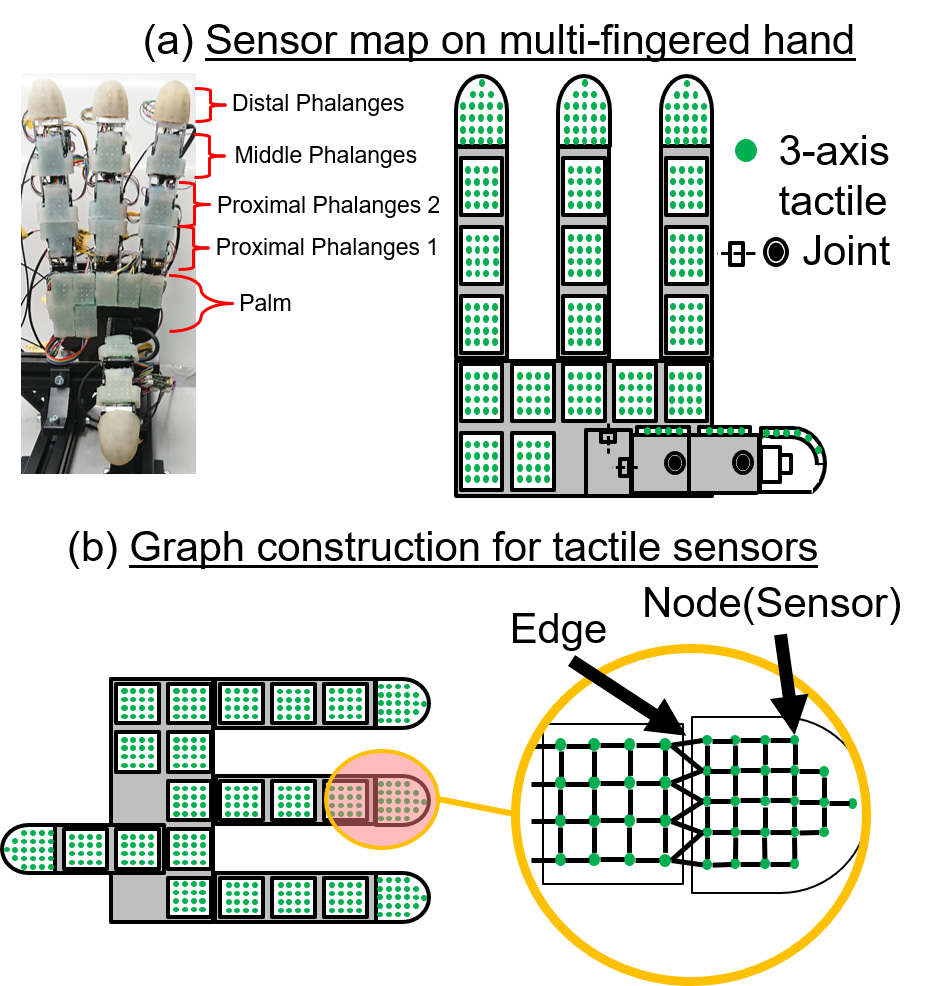}
      \caption{Tactile sensor alignments and its graph structure. (a) shows that uSkin sensors are mounted on the fingertips, phalanges and palm. (b) shows how we built the graph structure of tactile sensors. Each sensor chip is regarded as a node and they are connected by edges.}
      \label{sensormap}
   \end{figure}

Each sensor point of the uSkin as a node is connected by an edge as a graph structure (Fig. \ref{sensormap}). By constructing the information of each node and edge together, the grasping state of an object on the entire hand can be learned spatially. 
The tactile information is input to the GCN as follows:

\begin{eqnarray}
H^{n+1} = f(H^{n}, A)
\end{eqnarray}

where $f(H^{n}, A)$ is an output of the $n$-th graph convolution layer with inputs as $H^{n}$ and an adjacency matrix $A$ for the graph structure of the uSkin sensors on the Allegro Hand. The output becomes the $n+1$-th graph convolution layer's input $H^{n+1}$. Therefore, $H^{0}$ is the tactile information of the input layer in this study and $H^{N}$ is the last ($N$-th) graph convolution layer's input.

At this rate, only the features of neighboring nodes are used and the multiplication of the adjacency matrix $A$ introduces huge changes in the scale of the features.
To prevent those two problems, $f(H^{n}, A)$ is defined as follows:

\begin{eqnarray}
f(H^{n}, A) = \sigma(\hat{D}^{-\frac{1}{2}}\hat{A}\hat{D}^{-\frac{1}{2}}H^{n}W^{n})
\end{eqnarray}

where $\hat{A}$ includes the $A$ and an identity matrix $I$ to consider not only neighboring nodes but also the target node itself.  A symmetric normalization $\hat{D}^{-\frac{1}{2}}\hat{A}\hat{D}^{-\frac{1}{2}}H^{n}$ is introduced to prevent the scale change of the features. $W^{n}$ is the weight matrix of the $n$-th graph convolution layer and $\sigma$ is an activation function. Finally, tactile features are acquired and input to a fully-connected layer with other sensor information and the object property labels described in Section III-C.

\subsection{Motion Generation with Property Labels}
A model schematic of the motion generator is shown in Fig. \ref{schematic}. When the hand starts a manipulation from the initial grasping posture, joint and tactile sensor information are provided to the GCN. The tactile sensor information is input to the first convolution layer. Then, the features are obtained as the output from the last convolution layer. A variety of objects makes the single multi-fingered manipulation motion diverse due to their physical properties such as size, shape, softness, heaviness and slipperiness. Therefore, object property labels are prepared. Finally, the tactile features, the joint angles and the property labels are input to the first fully-connected layer. A next timestep of the joint angles is output from the output layer to adjust the posture of the fingers. By repeating these series of generation, the final grasping posture is reached. After a fixed number of time-steps the movement is stopped.


\section{Experiment Design}
\subsection{Training Data}

   \begin{figure*}[t]
      \centering
      \includegraphics[scale=0.6]{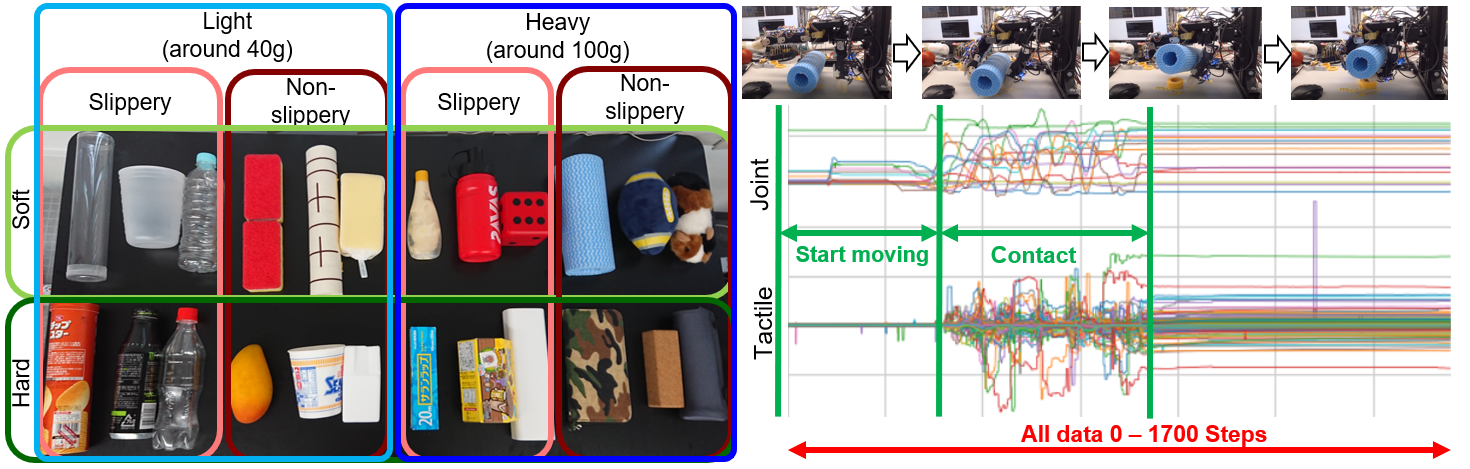}
      \caption{{\bf Target objects and data structure.} Left side: objects used for training data are shown as the first object on the left side of each group. The other two objects in each group are used as untrained objects. Right top row: An example of target manipulation motion is shown with kitchen paper. The motion is from precision to power grasps starting from a desk. A dataglove was used to generate natural training motions. Right bottom row: joint and tactile trajectories are shown. In the first several hundreds time steps, only joint trajectories move as fingers do not touch an object, yet. Afterwards, tactile measurement drastically alter. After reaching a power grasp (the final grasping posture), those trajectories do not change.}
      \label{datasetting}
   \end{figure*}

In the evaluation experiment of this study, we selected the motion of picking an object from a desk and holding it (i.e. a motion from precision to power grasps) as the target motion, see Fig. \ref{datasetting}.  
We selected this task as moving from precision to power grasps is a common strategy when humans pick up items, but has been seldom investigated for the following reasons. 
(1) it includes contact with the whole hand. Most robotic hands have no tactile sensors other than in the fingertips.
(2) various forces act between the object and the hand in 3 axes during the target in-hand manipulation.
(3) as the object is hidden inside the fingers during the in-hand manipulation, visual cameras are difficult to use because of occlusions, and hence tactile information is more important.

In this study, in order to realize manipulation of objects with various properties, various daily objects were chosen. Moreover, we focused on the three object properties: heaviness, hardness, and slipperiness, as they are usually embraced in tactile information. We prepared eight objects (2 heaviness x 2 hardness x 2 slipperiness) in total. 
As shown in Fig. \ref{datasetting}, a plastic tube (light, soft and slippery), a sponge (light, soft and non-slippery), mayonnaise (heavy, soft, and slippery), kitchen paper (heavy, soft, and non-slippery), a potato chip cylinder (light, hard, and slippery), a replica of a mango (light, hard, and non-slippery), saran wrap (heavy, hard, and slippery) and a purse (heavy, hard, and non-slippery) were used. Preliminary, we tried to use one-hot vectors (similar to \cite{onehotsergey}) to classify each object but the target manipulation failed as the vectors included many 0 values and the useful information for the manipulation was seemingly lost.

For the data collection, we remotely controlled our Allegro Hand using a CyberGlove (22-sensor model) from CyberGlove Systems, and acquired training data. Each training data is recorded every 10ms for 17 seconds. For each object, 10 successful trials of the target motion were conducted, and a total of 80 trials of training data were collected. Each object was put in a random pose below the Allegro hand, see Fig. \ref{datasetting}. However, the cylindrical objects were roughly aligned with the robot hand. Note that the data was collected with only visual feedback from the experimenter's eyes. The grasping states during the manipulation were roughly confirmed, for example, for the plastic tube it was visually checked that it did not deform during the manipulation. 
It should also be noted that we explored the possibility of tactile gloves with tactile feedback to improve our data collection method. However, for example, the HaptX system provides state of the art tactile feedback but only limited information about the user's hand configuration, and was therefore not suitable for our study.
We also collected larger sized datasets with 8 objects and 20 trials, and 16 objects and 10 trials. However, the model could not achieve the successful manipulation with the large datasets which include more diverse motions.

Each recorded trial was pre-processed before being input to the GCN. First, in each recorded trial, 
the datastream of the part where the finger is not moving immediately after the start of recording and the part where the finger is not moving after the end of manipulation are cut. Then, label information corresponding to the object was appended to the recorded data. Specifically, we prepare six labels in the following order: light, heavy, hard, soft, non-slippery, and slippery, and filled in 1 for each label when the object was defined with the properties, otherwise 0. For example, for a heavy, soft, and not slippery object (kitchen paper), the labels would be [0,1,0,1,1,0]. After this, smoothing was performed on the training data by taking the average of the data for 10 time steps (i.e. 5 time steps before and after the target time step) to reduce the noise of the data. Downsampling was performed to align the number of time steps recorded in the training data. The number of time steps used in this experiment was 330 for each trial and the total number of time steps was 26,300. The number of time steps used for training was 18,410, and for validation 7,890.

\subsection{Neural Network Settings}
\renewcommand{\arraystretch}{1}
\begin{table}[t]
\caption{Setting of Neural Networks}
\label{tab:1}
\centering
\tabcolsep = 1.0pt
\begin{tabular}{c|cccc}
\hhline{=====}
                                                                            & Model I                                                                                     & Model II                                                                                    & Model III                                                                                   & Model IV                                                                                                    \\ \hline
Network                                                                     & GCN                                                                                         & GCN                                                                                         & GCN                                                                                         & MLP                                                                                                         \\ \hline
\begin{tabular}[c]{@{}c@{}}Conv\\ layer\end{tabular}                        & \begin{tabular}[c]{@{}c@{}}(14,28,56,\\ 112,112,112)\end{tabular}                           & \begin{tabular}[c]{@{}c@{}}(14,28,\\ 56,112)\end{tabular}                                   & \begin{tabular}[c]{@{}c@{}}(14,28,\\ 56)\end{tabular}                                       & -                                                                                                           \\ \hline
\begin{tabular}[c]{@{}c@{}}Input \\ to Conv\end{tabular}                    & \multicolumn{3}{c}{\begin{tabular}[c]{@{}c@{}}Tactile (384 sensor \\ chips * 3 axes)\end{tabular}}                                                                                                                                                                                 & -                                                                                                           \\ \hline
\begin{tabular}[c]{@{}c@{}}FC\\ layer\end{tabular}                          & \multicolumn{3}{c}{\begin{tabular}[c]{@{}c@{}}(8000,1000,\\ 120,50)\end{tabular}}                                                                                                                                                                                                      & \begin{tabular}[c]{@{}c@{}}(1500,3000,1500,\\ 700,350,100,50)\end{tabular}                                  \\ \hline
\multicolumn{1}{l|}{\begin{tabular}[c]{@{}l@{}}Input \\ to FC\end{tabular}} & \begin{tabular}[c]{@{}c@{}}Tactile feature\\  (45472),\\ Joint (16), \\ Label (6)\end{tabular} & \begin{tabular}[c]{@{}c@{}}Tactile feature\\  (45472),\\ Joint (16), \\ Label (6)\end{tabular} & \begin{tabular}[c]{@{}c@{}}Tactile feature\\  (22736),\\ Joint (16), \\ Label (6)\end{tabular} & \begin{tabular}[c]{@{}c@{}}Tactile (384 sensor \\ chips * 3 axes),\\ Joint (16), Label (6)\end{tabular} \\ \hline
\begin{tabular}[c]{@{}c@{}}FC \\ layer\end{tabular}                         & \multicolumn{3}{c}{\begin{tabular}[c]{@{}c@{}}(8000,1000,\\ 120,50)\end{tabular}}                                                                                                                                                                                                      & \begin{tabular}[c]{@{}c@{}}(1500,3000,1500,\\ 700,350,100,50)\end{tabular}                                  \\ \hline
Output                                                                      & \multicolumn{4}{c}{Joint (16)}                                                                                                                                                         \\ \hline
\hhline{=====}
\end{tabular}
\end{table}
In order to verify the effectiveness of the GCN, we conducted a comparison experiment by changing the number of graph convolution layers. To do this, 4 neural network models were prepared, as shown in Table \ref{tab:1}. Model I was a GCN, which had 6 graph convolution layers with a Conv layer size of [14, 28, 56, 112, 112, 112] and four fully-connected layers (FC layer) with sizes of [8000, 1000, 120, 50]. Model II was also a GCN which had 4 graph convolution layers with a Conv layer size of [14, 28, 56, 112] and the same fully-connected layers as Model I. Model III was also a GCN which had 3 graph convolution layers with a Conv layer size of [14, 28, 56] and the same fully-connected layers as Model I. 

In addition, a multi-layer perceptron (MLP) was prepared as Model IV to confirm whether the convolution layers were necessary or not. The MLP setting was as follows: the total number of fully-connected layers was 7, and the size of each layer was [1500, 3000, 1500, 700, 350, 100, 50]. Note that 3 axes shown in `Input to Conv' of Table \ref{tab:1} are 3 channels for Model I, II and III, while 3 inputs in `Input to FC' for Model IV.

For all the networks, no optimization techniques such as pooling and dropout were used. The input for the networks were tactile, joint measurements and object property labels. The number of dimensions of the input is 4 (finger) x 4 (joint) = 16 dimensions for joint angle, 384 (sensor) x 3 (axis) = 1152 dimensions for tactile information, and 6 dimensions for object property, so the total number of dimensions of the input is 1174. The number of dimensions of the output is 4 (fingers) × 4 (joints) = 16 dimensions and the time step to be predicted is 10 after the current time step. The learning rate of the Adam optimizer is 0.00001. The batch size is 100 for both training and validation. The number of training epochs was 5,000. Relu was used as an activation function for all layers, including convolution and fully-connected (FC) layers, except for the output layer, which had no activation function. Adam was used as the optimizer for all architectures with a learning rate of 0.00001, step size of 0.0001, first exponential decay rate of 0.9, second exponential decay rate of 0.999, and small value for numerical stability of 1e-08. All the networks were built with the PyTorch and PyTorch Geometric libraries for Python 3 and trained with NVIDIA V100 SXM2 on AI Bridging Cloud Infrastructure (ABCI) provided by the National Institute of Advanced Industrial Science and Technology (AIST).

\section{Evaluation}
\subsection{Comparison of Graph Convolution Layers}
    \begin{figure}[t]
      \centering
      \includegraphics[scale=0.28]{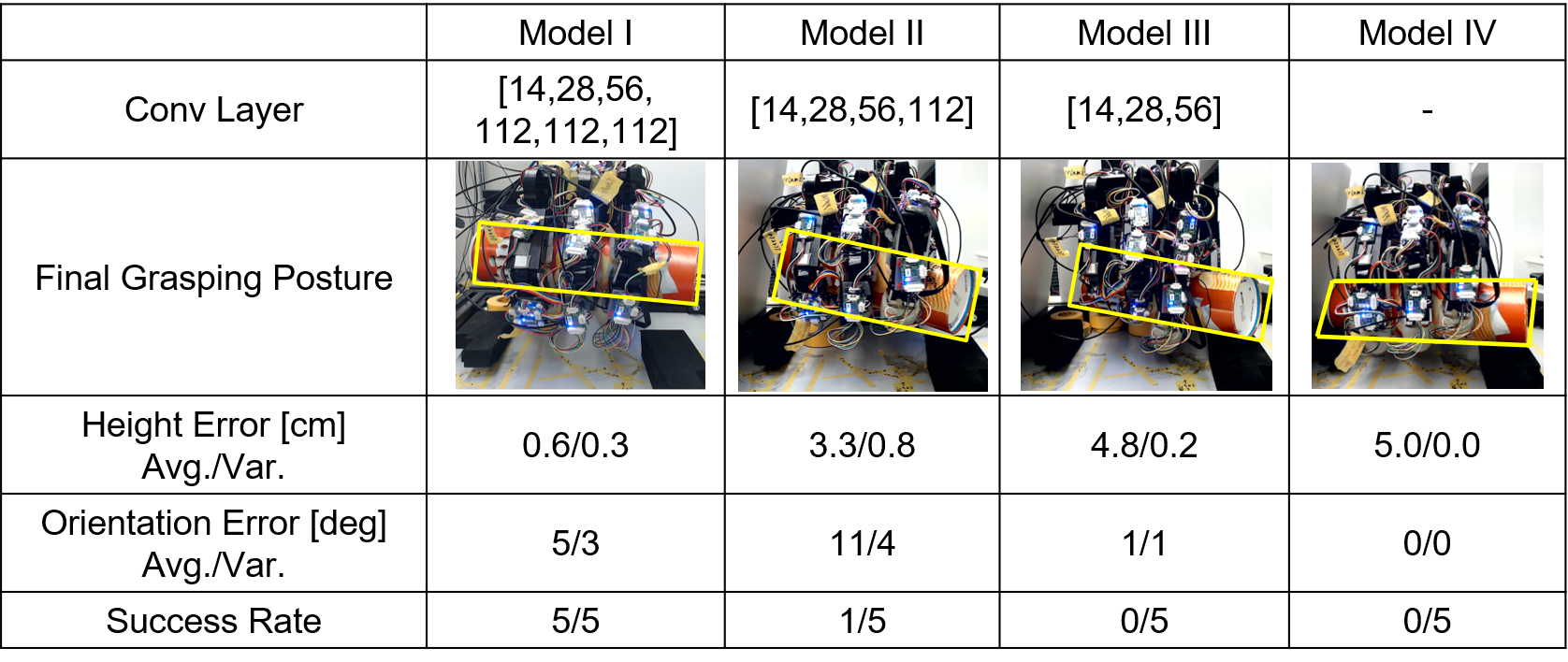}
      \caption{Comparison result of 4 network models. The difference is the number of Conv layers and Model IV has 3 more FC layers than the others. The yellow line shows a rough curvature of the potato chip cylinder so that height and orientation errors can be easily confirmed.}
      \label{result}
  \end{figure}
  \begin{figure}[b]
      \centering
      \includegraphics[scale=0.32]{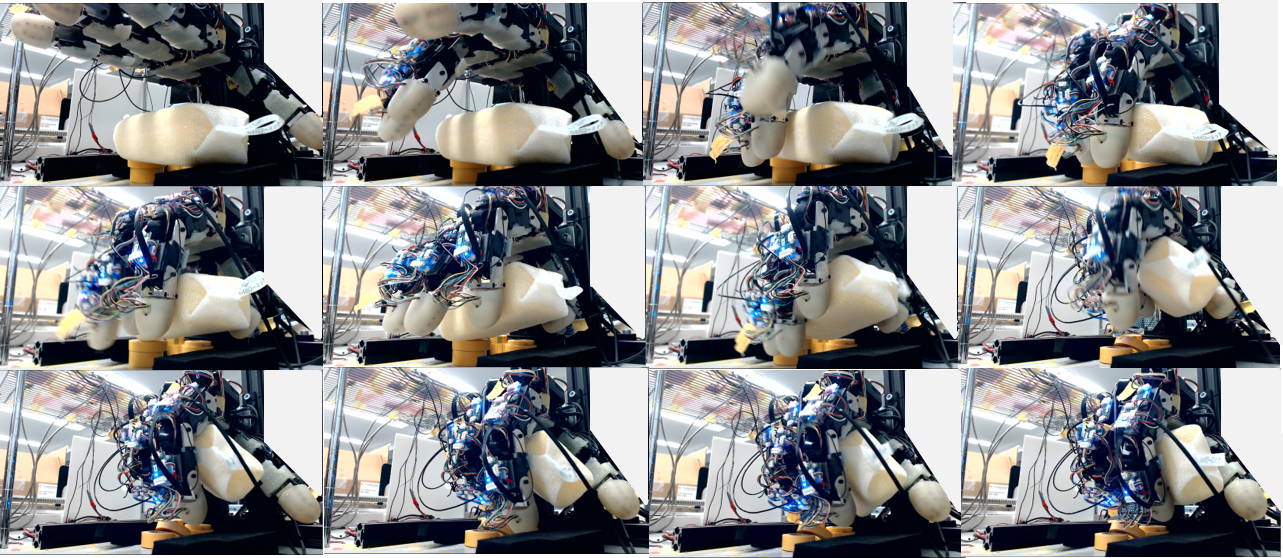}
      \caption{Successful in-hand manipulation with a sponge. In the initial grasping posture, the hand open its fingers. The fingers gradually get close to the object. After touching the object with the fingertips, the motion is generated to pick it up. The motion embraces finger gaiting and rolling.}
      \label{manipulation}
  \end{figure}

The four models mentioned in Section IV-B were used. The definition of success is whether the distance between the palm of the Allegro Hand and the target object is under 2 cm and whether the orientation of the object to the palm is under 15 degrees at the final grasping posture. The manipulation with each model was conducted five times. A potato chip cylinder was used as the target object as its elongated shape made it easy to measure the height and orientation, while some other trained objects showed similar results in preliminary experiments to the case of the potato chip cylinder. 

The result is shown in Fig. \ref{result} including an example of the final grasping posture generated by each model. Model I could achieve the highest success rate, 5 out of 5 times. Importantly, the manipulation was conducted with each finger cooperating with each other. Finger gaiting was often confirmed during the manipulation, for example when the index finger and thumb pinched the object, the middle finger touched the object and the index finger broke contact and touched the object again at a different contact position. Interestingly, some fingers moved in synchrony and lifted and spun the object dynamically, using the friction of its silicone skin to reach the final grasping posture. 
Please refer to the video for an example of such a motion. 
An example of successful manipulation is shown in Fig. \ref{manipulation}. 
On the other hand, Model II which has a lower number of Conv layers achieved the manipulation only 1 out of 5 times. It succeeded in pinching the object (precision grasp) but could not reach the final grasping posture (power grasp) with the distance error under 2 cm. Similar manipulation motions were generated by Model III but 0 out of 5 trials were successful. Model IV, which is the MLP, produced the worst motions and could not even pinch the object up at all. That is why the height error was 5 cm with 0 variance and the orientation error was 0 degree with 0 variance. From this result, having sufficient number of graph convolution layers enables 
dexterous manipulation 
with coordinated finger movements.

\subsection{Graph Structure and Robotic Configuration}
  \begin{figure}[t]
      \centering
      \includegraphics[scale=0.42]{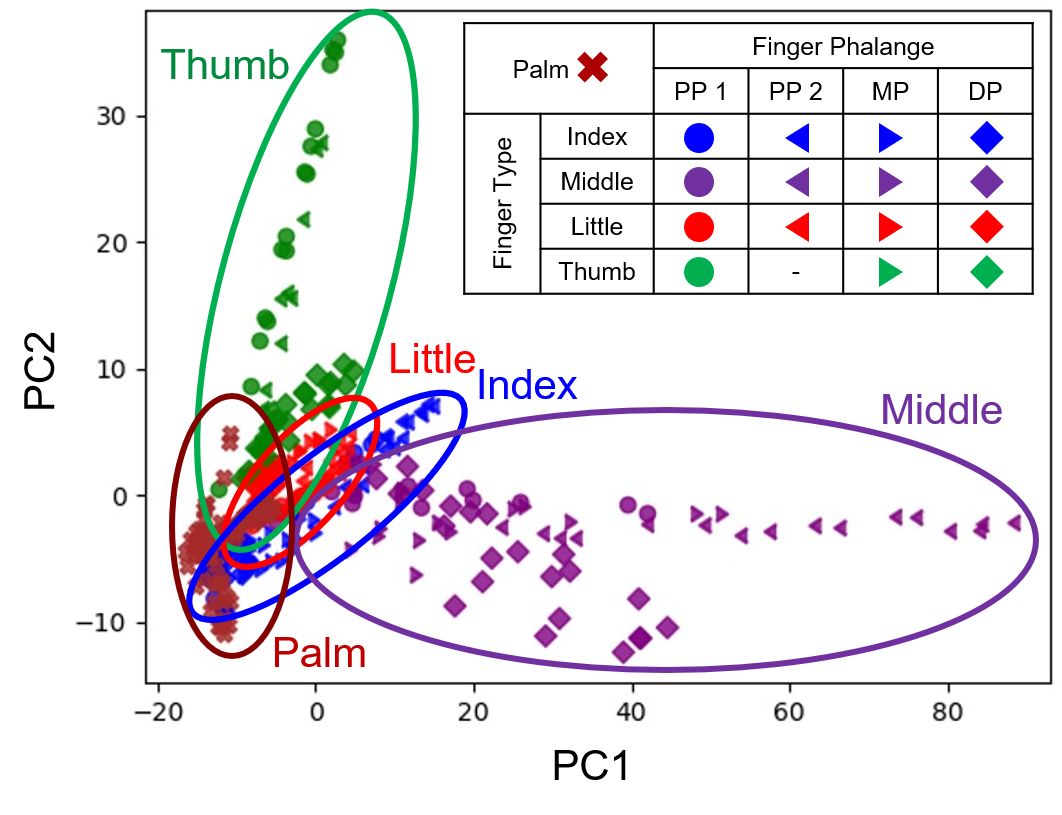}
      \caption{A PCA map for the features generated by nodes in the GCN during the final grasping postures. Each color shows each segment on the Allegro Hand (i.e. green for thumb, red for little finger, blue for index finger, purple for middle finger and brown for palm). This PCA map is made of PC1 and PC2 axes. Dots and triangles represent phalanges, and diamonds fingertips. Cross marks represent the palm. PP 1 and PP 2 represent the lower and upper part of the proximal phalanges, respectively. MP is the middle phalange, DP is the distal phalange. Those are depicted in Fig. \ref{sensormap}. A cluster for the palm is placed under all fingers. The clusters for the fingers are next to each other. It indicates that the GCN extracts a functional representation of the robotic configuration. 
      For the thumb and middle finger, the clusters are bigger.
      Coincidentally, we could also confirm that the tactile measurements in the last time step of our manipulation are larger for the thumb and middle finger than for the other tactile sensors.
      Those fingers mainly support a grasped object against gravity during the final grasping postures.}
      \label{nodepca}
  \end{figure}

As described in Section V-A, there was a large difference in success rates of in-hand manipulation among the networks. In order to investigate the factors behind the successful manipulation, a principal component analysis (PCA) was used and the tactile features obtained from the last graph convolution layer were studied for Model I. Only features obtained from the time steps when the manipulation of an object was completed, after reaching the final grasping posture, were used. The time steps were sampled from training data of all the objects. Then, dimensional compression was performed in the direction of the time steps by the PCA so that node features of the GCN were extracted. The shape of the compressed features is [384 (nodes), 2 (principal components)]. The 2 PCs were produced from the features by compressing 45 (time steps) × 8 (objects) × 10 (trials) × 112 (filters in the last convolution layer) dimensions.

As shown in Fig. \ref{nodepca}, clusters emerge for each finger and each part of the finger, i.e. the fingertips, the finger phalanges, and the palm. The cluster of the palm was located below all fingers. The spatial and functional relationship between the fingers is extracted by the features. For example, index, middle and little fingers scatter next to each other as their role during manipulation is similar, while the thumb and middle fingers are symmetrically scattered as they are opposing fingers to each other.

Moreover, the tactile information seems to affect the clusters. The thumb and middle finger are widely distributed. We assume this is because of larger importance of the tactile information from those fingers than those from the other fingers as the thumb and middle finger support a grasped object against gravity during the final grasping postures. On the other hand, the little finger is distributed small because tactile information is often small as the finger had less chances to touch the grasped object due to the size or shape of the object. Note that those clusters did not emerge before training, unlike to \cite{gcn} which mentioned that a GCN before training already produced clusters. The clusters in this study were produced by Model I after 5,000 training epochs (i.e., Model I). In addition, the feature map with clear clusters was acquired only from the last graph convolution layer not from the rest of the layers. Overall, learning manipulation motions was meaningful for acquiring geodesical or robotic features with a sufficient number of convolutions.


\subsection{Analysis on Property Labels and Touch States}
  \begin{figure}[t]
      \centering
      \includegraphics[scale=0.5]{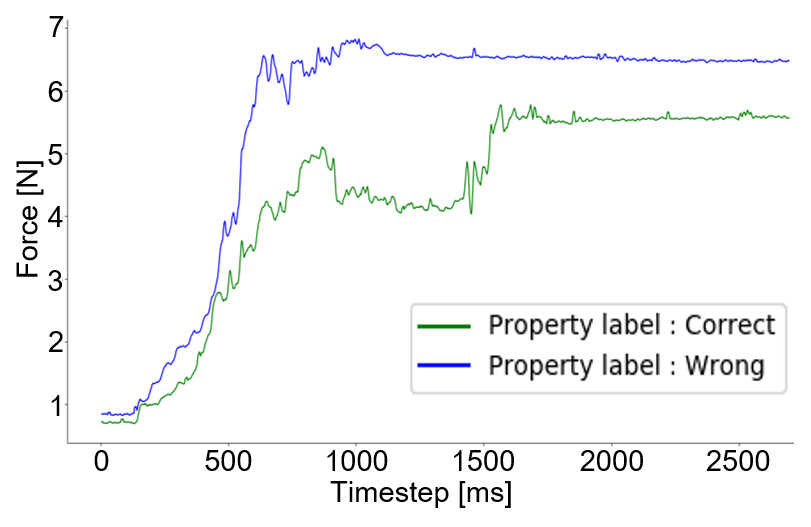}
      \caption{Total grasping forces during in-hand manipulation. Trajectories of grasping forces on a plastic tube are shown. Green and blue lines show the forces generated by Model I given correct and wrong labels, respectively.
      }
      \label{graspstate}
  \end{figure}
  \begin{figure}[t]
      \centering
      \includegraphics[scale=0.35]{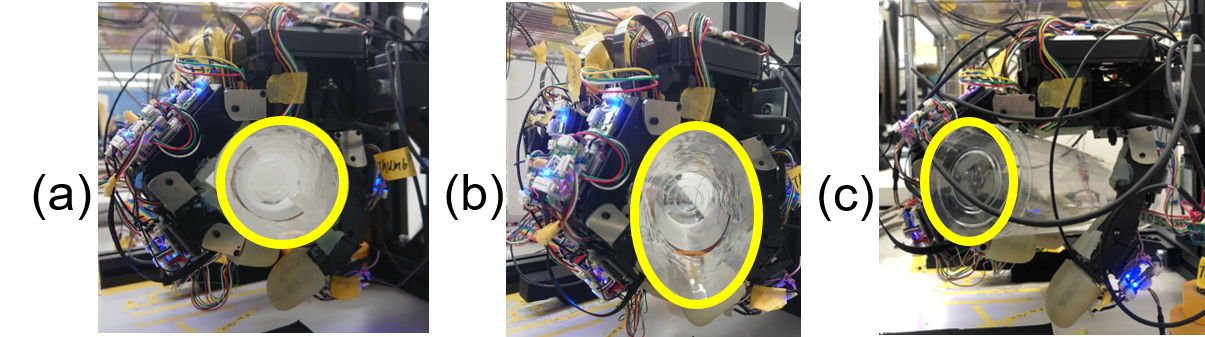}
      \caption{Grasping states with a soft object. (a) shows the final grasping posture when Model I is given the correct labels. The grasped object is not deformed from its side. (b) shows the final grasping posture with Model I given the wrong labels. The grasped object is deformed into an oval shape from its side. (c) shows the final grasping posture with Model I trained with no labels. The object is in a loose grasp and the manipulation was never successful.}
      \label{deformation}
  \end{figure}
  
To confirm the effectiveness of using property labels, we performed manipulation by changing the property labels for in-hand manipulation. In this comparison experiment, the property labels for a target object were specified and input to the GCN (Model I) with a soft plastic tube as the target object. Two sets of property labels were used as input: correct labels and incorrect labels. The correct label consists of light, soft, and slippery, which are the properties of the object, while the wrong label consists of heavy, hard, and slippery which is actually for saran wrap. The GCN (Model I) trained with no labels was also prepared for this study. However, it never succeeded in performing the manipulation and produced different motions in every trial (e.g. pressing an object to the desk and extending some fingers in the final grasping posture.) It seems that the model could not learn to generate the desired motion but a random one without property labels (Fig. \ref{deformation}-(c) shows one example).

As shown in Fig. \ref{graspstate}, the total force from all tactile sensors was always higher when the wrong property label was used than when the correct property label was used. The result with no property labels is not shown in Fig. \ref{graspstate} as each trial showed random motions.

Fig. \ref{deformation} shows the cross-section of the object in the final grasping posture when the correct property label (Fig. \ref{deformation}-(a)) and the incorrect property label (Fig. \ref{deformation}-(b)) are used. When the correct property label was used, the soft plastic cylinder was not squeezed, and the cross section was circular as shown by the yellow line. On the other hand, when the wrong property label was used, the soft plastic cylinder was crushed, and the cross section was deformed into an oval as shown by the yellow line.

From this result, when the correct label was used, the hand could perform the manipulation with an appropriate grasping force, and therefore, did not crush the object. It was proven that one GCN learns many multi-fingered manipulation motions which are variable due to object properties at the same time and the property labels could adjust the motions.

\subsection{Robustness Test against Disturbance}
  \begin{figure}[t]
      \centering
      \includegraphics[scale=0.28]{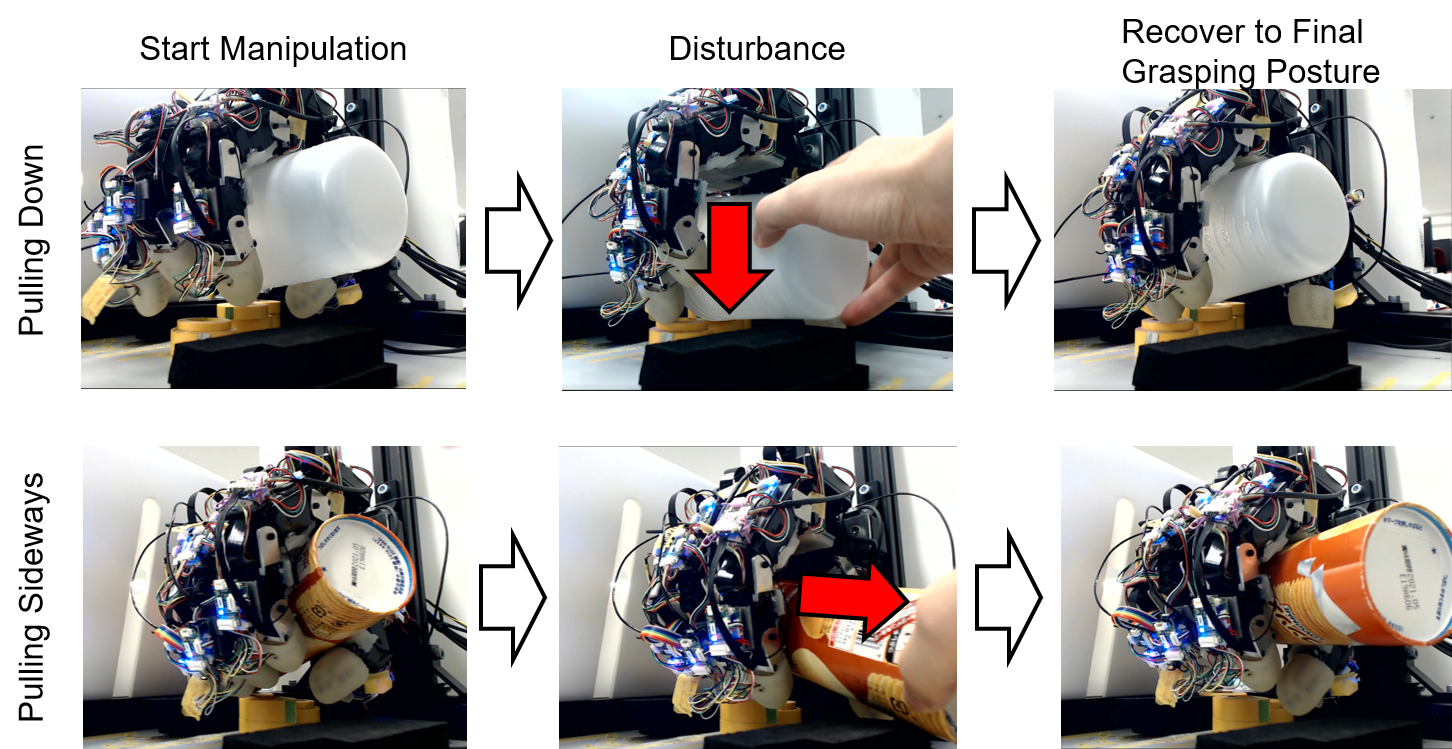}
      \caption{Examples of recovery from a huge disturbance. From the left side, the hand starts to manipulate an object. In the middle, an experimenter pulled down or sideways the object as a disturbance; the experimenter felt a large friction from the grasping force of the hand. Red arrows show the direction of pulling by the experimenter. On the right side, the hand deals with the disturbance and reaches the final grasping posture.
      }
      \label{disturbance}
  \end{figure}
  
For the evaluation of robustness of the proposed method, an experimenter pulled down or sideways on a grasped object during manipulation conducted by the GCN (Model I) shown in Fig.~\ref{disturbance}. Even though a huge disturbance which totally changes the grasping state and position of the object happened, the hand could recover and reached the final grasping posture successfully. What the model did was to predict joint angles of a next time step from a current grasping state (joint angles and tactile information). The interesting point is when pulling down disturbance happens for example, the state of joint angles may be similar to the ones for the final grasping posture while the state of tactile information may be similar to the one for initial grasping state (the object is touched at fingertips). This situation never occurred during training, but interestingly the network could handle this situation as well. It showed the robustness of the proposed model. We expect that the widely distributed tactile sensors are beneficial, as they ensure that they still contact the object after the disturbance.

\subsection{Generalization to Untrained Objects}
\begin{table*}[t]
\centering
\caption{Achievement of the Manipulation with Untrained Objects}
\renewcommand{\arraystretch}{1}
\begin{tabular}{c|cc|cc|cccc|c}
\hhline{==========}
\multirow{1}{*}{}     & \multicolumn{4}{c|}{\begin{tabular}[c]{@{}c@{}}Light (around 40g)\end{tabular}} & \multicolumn{4}{c|}{\begin{tabular}[c]{@{}c@{}}Heavy (around 100g)\end{tabular}}     & \begin{tabular}[c]{@{}c@{}}Success Rate\end{tabular} \\ \cline{2-10} 
                      & \multicolumn{2}{c|}{Slippery}           & \multicolumn{2}{c|}{Non-Slippery}        & \multicolumn{2}{c|}{Slippery}                      & \multicolumn{2}{c|}{Non-Slippery} &                                                                \\ \cline{1-9}
\multirow{2}{*}{Soft} & Plastic beaker          & 5/5          & Paper scroll              & 3/5          & Water bottle          & \multicolumn{1}{c|}{3/5}  & Stuffed foot ball toy    & 4/5    &                                                                \\
                      & PET bottle              & 3/5          & Sponge                    & 4/5          & Stuffed dice toy      & \multicolumn{1}{c|}{4/5}  & Stuffed hamster toy      & 3/5    & 59/80                                                          \\ \cline{1-9}
\multirow{2}{*}{Hard} & Canned drink            & 4/5          & Cup noodle                & 4/5          & Snack package         & \multicolumn{1}{c|}{5/5}  & Cork block               & 3/5    & 73.75\%                                                        \\
                      & PET bottle              & 4/5          & Polystyrene case          & 3/5          & Cleaning brush case   & \multicolumn{1}{c|}{3/5}  & Bottle case              & 3/5    &             
\\
\hhline{==========}
\end{tabular}   
\label{tab:2}
\end{table*}

Finally, Model I was applied for multi-fingered manipulation of untrained dafily objects. Sixteen objects were prepared and each group had 2 objects, shown in Fig. \ref{datasetting}. When conducting the manipulation experiment, Model I was input object property labels for each untrained object according to the allocated groups. The success definition is the same as the one in Section V-A. An initial grasping position for each object was roughly set where the hand can succeed the manipulation, within a range of around 1 cm. Although the untrained objects had a variety of physical properties (e.g. size, shape, texture, weight, etc...), all the untrained objects were chosen based on property groups (heaviness, softness and slipperiness). 

As shown in Table \ref{tab:2}, a lot of objects were successfully manipulated many times. Most of them were dynamically manipulated using skills such as finger-gaiting or rolling-contact, similar to the ones in Section V-A, despite the many differences in the physical properties of the untrained objects. Noteworthy skillful manipulation was achieved with a plastic beaker: the object was tilted over 45 degrees and looked difficult to recover during pinching with the thumb and index finger, but the middle and little finger supported the other fingers and finger-gaiting was conducted resulting in reaching the final grasping posture. Moreover, the manipulation was not redundantly conducted, for example, when a stuffed dice toy was manipulated, finger-gaiting was not conducted as often as for other objects because the dice was large but still the final grasping posture was achieved. This indicates that our model sufficiently learned the goal grasping state. Overall, 73.85 \% was achieved as the total success rate of manipulation with all the untrained objects. 

As a limitation of our approach it should be mentioned that the initial grasping position has to be within a small area to enable successful manipulation. Also the gap between the finger segments was problematic as objects sometimes got stuck in the gap. Autonomously acquiring property information would increase the generality of our method.

\section{Conclusions}

This study presented a control method of multi-fingered manipulation with a variety of object properties. A GCN acquiring tactile and geodesical features of a robot hand and achieved dexterous in-hand manipulation with synchronized finger movements. Furthermore, labels for each object property enabled the GCN to change manipulating motions depending on the target object. Specifically, grasping forces were reduced and the soft object was not crushed and successfully manipulated. Moreover, it was confirmed that each node of the GCN clusters follows the robotic configuration, the role of each finger and the tactile information. Finally, the manipulation under an experimenter's disturbances and the manipulation with untrained objects were robustly achieved. 
Overall we could show that intrinsic object features are crucial for object manipulation. 

As future work, visual cameras will be used for locating the position of target objects to increase the success rates. Also, the gap between the finger phalanges should be reduced, for example with a mechanical design such as \cite{jointgap}. Moreover, the object properties should be automatically acquired in real-time but not prepared as labels for more autonomous control. Initial experiments to this end have already been conducted as we tried object property recognition with a GCN in real-time, but further improvements are needed.
Object picking with a robot arm is a next step as the hand achieved object picking from a desk in this study.

\section*{Acknowledgement}
{This research was supported by the Japan Science and Technology Agency, ACT-I Information and Future Acceleration Phase with a grant number of JPMJPR18UP and Moonshot R\&D with a grant number of JPMJMS2031.}

\bibliographystyle{IEEEtran}
\bibliography{reference}
\end{document}